\documentclass[conference]{IEEEtran}
\IEEEoverridecommandlockouts
\usepackage{cite}
\usepackage{amsmath,amssymb,amsfonts}
\usepackage{algorithmic}
\usepackage{graphicx}
\usepackage{textcomp}
\usepackage{xcolor}
\usepackage{stmaryrd}
\usepackage{booktabs}
\usepackage{subcaption} 
\usepackage{url}
\usepackage{hyperref}

\usepackage{xcolor}
\usepackage{tikz}
\usetikzlibrary{shapes,arrows,positioning}
\def\BibTeX{{\rm B\kern-.05em{\sc i\kern-.025em b}\kern-.08em
    T\kern-.1667em\lower.7ex\hbox{E}\kern-.125emX}}
\begin{document}
\definecolor{darkgreen}{RGB}{34, 85, 34}
%%% BLOC DIAGRAM 
\tikzstyle{block} = [draw, rectangle, minimum height=2.2em, minimum width=4em, text width=4em, align=center, font=\scriptsize]
\title{\LARGE \bf {A Sim-to-Real Integration Pipeline for Training and Deployment\\
of Chunk-Based VLA Manipulation Policies}}

\author{
\IEEEauthorblockN{\small
Mathilde Kappel$^{*}$, Clémence Grislain$^{*}$, Mohamed Chetouani, Olivier Sigaud,\\
Louis Annabi, Faïz Ben Amar, Stéphane Doncieux, Mahdi Khoramshahi}
\IEEEauthorblockA{\small
Institute of Intelligent Systems and Robotics (ISIR), CNRS, Sorbonne University, Paris, F-75005, France\\[2pt]
{\scriptsize \textit{$^{*}$Equal contribution}}}
}
\maketitle
\thispagestyle{plain}
\pagestyle{plain}

\vspace{-1em}
\begin{abstract}
Vision-Language-Action (VLA) models have become a prominent paradigm for mapping multimodal inputs, including semantic instructions, visual observations of the scene, and proprioceptive observations, to robot actions. Most state-of-the-art models predict actions in the end-effector pose space as sequences of action chunks. Training and evaluating these models requires large-scale collections of real-world demonstrations, pairing robot actions with the corresponding visual and proprioceptive observations. Collecting such data on real hardware typically relies on human teleoperation, making the process costly, time-consuming, and difficult to scale. We present an open-source sim-to-real experimental protocol that addresses this bottleneck: expert trajectories generated in simulation are replayed open-loop on a real Franka FR3 setup, where the corresponding real visual and proprioceptive observations are recorded and converted into a format compatible with VLA training. The same deployment stack is then reused, in closed-loop, to evaluate a trained policy on that setup, so that data collection and evaluation share an identical hardware configuration. Because each real recording is paired with the simulated trajectory that produced it, the protocol also yields a direct measurement of the sim-to-real gap. We release the collected datasets on Hugging Face together with the pipeline source code \url{https://gitlab.isir.upmc.fr/kappel/sim2real_public_chunk_control/}.
\end{abstract}

% \begin{IEEEkeywords}
% component, formatting, style, styling, insert
% \end{IEEEkeywords}

\section{Introduction}

Vision-Language-Action (VLA) models map a language instruction and visual observations to robot actions, and have recently shown strong generalization across tasks and embodiments~\cite{octo, openvla, pi0}. Most of them predict \emph{chunks} of end-effector displacements~\cite{diffusion_policy}, which
makes their action space largely agnostic to the underlying controller. This property is what makes such policies natural candidates for sim-to-real transfer: a chunk produced in simulation is, in principle, directly executable on a real arm.

In practice, two obstacles remain. First, real-world demonstrations are expensive: most VLA datasets rely on human teleoperation~\cite{droid}, which does not scale and couples data quality to operator skill. Second, the sim-to-real gap is rarely \emph{quantified}. Pipelines are typically validated by an end-task success rate, which conflates policy quality, perception, and
physical mismatch into a single number, and gives no indication of where the mismatch originates.

This paper presents an open-source experimental framework for scalable real-world data collection and systematic quantification of the sim-to-real gap. Expert trajectories are generated in simulation and replayed open-loop on a real Franka FR3, while the real visual and proprioceptive observations are recorded. The result is a real-world dataset whose action labels come from a simulated expert, at no teleoperation cost. Because every real recording is paired with the simulated trajectory that produced it, the same protocol turns the real robot into a measurement instrument: the
deviation between the two trajectories is a direct, physically grounded estimate of the sim-to-real gap, independent of any learned policy. The deployment stack is designed so that the exact same components are reused, in closed-loop, when a trained policy is evaluated on the setup.
Our contributions are:
\begin{itemize}
    \item an open-source sim-to-real pipeline for chunk-based VLA policies, covering open-loop expert replay for data collection and closed-loop execution for policy evaluation;
    \item dataset of $200$ paired sim-to-real cube-pushing trajectories released on Hugging Face, together with a policy-independent characterization of the sim-to-real gap.
\end{itemize}

\section{Related work}

\textbf{Vision-language-action policies.} Large-scale imitation learning has produced policies that condition on language and images to output low-level actions~\cite{openvla, octo, pi0}. A common design choice, inherited from previous work (Diffusion Policy \cite{diffusion_policy}), is to predict a chunk of k future actions rather than a single
step, which reduces compounding error and inference latency. Our pipeline targets this family: it assumes only that the policy emits chunks of relative end-effector displacements and a gripper state, and is otherwise agnostic to the model architecture.

\textbf{Data collection and deployment pipelines.} Most real-world VLA datasets are built by teleoperation~\cite{droid}, and open-source frameworks such as LeRobot~\cite{lerobot} standardize their format and the deployment loop. Simulation benchmarks such as CALVIN~\cite{calvin} provide language-conditioned expert trajectories at scale, but without a real-robot counterpart. Our protocol bridges the two by replaying simulated experts on hardware, which yields real observations with simulation-grade action labels. Unlike approaches that evaluate the gap end-to-end, through the difference in task success between simulation and hardware, we replay a simulated command sequence on the real robot that yields a paired execution: the same demonstration run in both worlds. This pairing is what makes the discrepancy measurable at the trajectory level rather than summarized by a single policy-dependent aggregate score~\cite{aljalbout}.

\textbf{Measuring the sim-to-real gap.} The gap is most often quantified indirectly, through the drop in success rate between simulation and reality, or reduced through domain randomization~\cite{domain_rand} and real-to-sim system
identification~\cite{aljalbout}. Direct trajectory-level comparison is less common, in part because simulated and real executions are not temporally aligned. We rely on geometric distances between end-effector paths, the discrete Fréchet distance~\cite{frechet} and the mean point-to-curve distance, which provide a policy-independent measurement of the physical mismatch.

\section{Method}
Section~\ref{sec:setup} describes the paired simulated and real-world setup. Section~\ref{Chunking Learning Model Description} introduces the action chunk representation targeted by the pipeline, and how these chunks are processed and executed in the real robot. Section~\ref{subsec:open-loop} details the expert and inference deployment protocols. Finally, Section~\ref{ROS2 sim2real communication interface} presents the underlying \texttt{ROS2} interface.

\subsection{Sim2Real Setup}
\label{sec:setup}

\begin{figure}[!t]
        \includegraphics[width=1.\linewidth]{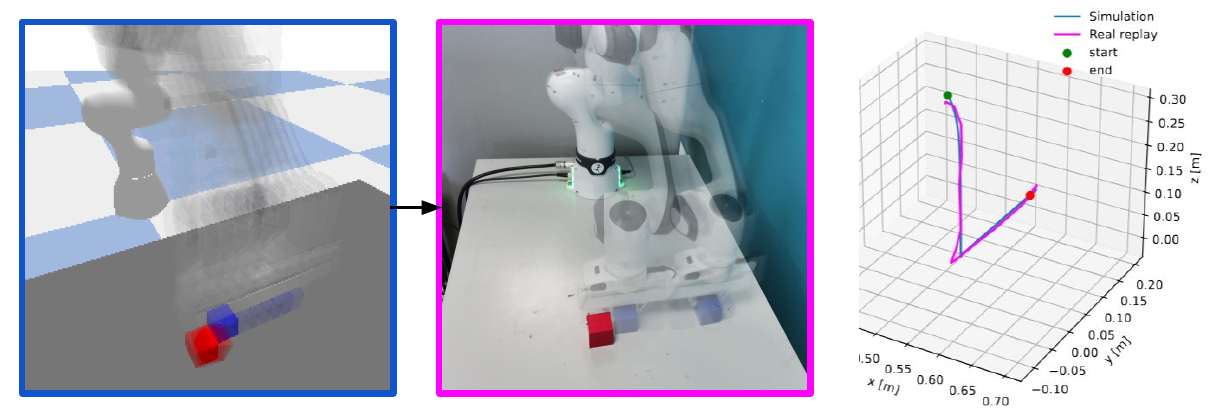}
        \label{fig:real_env}
    \caption{Visual observations of the paired environments from simulation (\textcolor{blue}{blue}) to real-world data collection (\textcolor{magenta}{magenta}).}
    \label{fig:sim2real_envs}
\end{figure}

In this work, we consider a paired simulated and real-world setup, composed of a Franka FR3 arm and a set of objects for which a URDF model is available, in our case, two colored cubes. The simulated environment is built on the PyBullet simulator \cite{pybullet} and mirrors the real scene: a Franka FR3 mounted on a table, with the world frame origin placed at the robot base in both worlds. Sharing the same origin is what allows simulated and recorded end-effector poses to be compared directly, without any pose transformation. Visual observations are provided by a fixed Orbec Femto Bolt RGB camera at $1000\times 1000$ resolution. The simulated camera and real-world camera are placed so that both worlds observe the same scene from a similar viewpoint. A precise calibration of the camera pose is not necessary, since the transfer operates directly on the real reference frame. Before each expert replay, the initial scene configuration is reproduced by using the robot itself as a Cartesian pointer: the operator moves the end-effector to the position where the simulated cube is located, and places the real cube accordingly, which bounds the initial object placement error to approximately 3.4 cm. Figure~\ref{fig:sim2real_envs} shows a visual observation from each environment. Several physical properties are left unmodeled, cube mass and inertia, table friction are not identified from the real setup, and constitute the main expected sources of the residual sim-to-real gap.

\subsection{Action Chunk Representation, Processing, and Execution} 
\label{Chunking Learning Model Description}

\textbf{Action Chunk Representation.} We consider a VLA policy $\pi_\text{VLA}$ that predicts action chunks of fixed size $k \in \mathbb{N}_+$.
Each action chunk $a_{SimChunk}^t$ is indexed by a discrete index $t \in \llbracket 1,m \rrbracket$. We denote $ o_t \in \mathbb{R}^{3\times H\times W}$ as an RGB visual observation of shape $H\times W$ of the real scene given to the VLA to generate the action chunk $a_{\mathrm{SimChunk}}^{t} \sim \pi_\text{VLA}\left(. \mid o_t\right)$. The deployment of a full task using $\pi_\text{VLA}$ consists of deploying $m$ action chunks on the real robot.  
Observations $o_t$ are obtained by the real-world camera.

\textbf{Action Chunk Processing.} \label{Integration of one action chunking}
At each timestep $t$ of the policy $\pi_\text{VLA}$ real-world deployment, results in the computation and execution of one action chunk $a_{SimChunk}^t = (a^t_{sim,1},a^t_{sim,2},....,a^t_{sim,k})$. This sequence embodies the next $k$ model-inferred actions $a^t_{sim,i}$ to be sent to the real robot. A model-inferred simulated command $a^i_{sim}$ corresponds to a variation of the end-effector frame and finger configuration to be sent to the real robot.

On the one hand, one variation $i$ of the end-effector frame is formalized as follows.
Let $t \in \llbracket 1,m \rrbracket$ and $i \in \llbracket 1,k \rrbracket$. We define
$\Delta \mathrm{eef}_i^t = (\Delta x_i^t,\Delta y_i^t,\Delta z_i^t,\delta q_{1,i}^t,\delta q_{2,i}^t,\delta q_{3,i}^t,\delta q_{4,i}^t) \in SE(3)$.
 with $\Delta$ being change in translation and $\delta $ change in the orientation of the end-effector.
On the other hand, the gripper configuration is formalized as a boolean $g \in \{0,1\}$ corresponding to opened or closed gripper. Let us define the action space $\mathcal{A} = SE(3) \times \{0,1\}$. One robot simulation command is formalized as the following tuple : 
\begin{equation}
    a^{t}_{sim,i}=  (\Delta \mathrm{eef}_i^t,g_i^t) \in \mathcal{A} .
\end{equation}
One $a_{SimChunk}^t$ accumulates $k$ sequential simulated command and has the following formalization:
\begin{equation}
    a_{SimChunk}^t
    =
    \left(
        (\Delta \mathrm{eef_1}^t,g_1^t),
        \ldots,
        (\Delta \mathrm{eef_k}^t,g_k^t)
    \right) \in   \mathcal{A}^k .
\end{equation}

For each timestep $t$, the computed action chunk $a_{SimChunk}^t$ is converted into a sequence of target states for the real robot's end-effector. To do so, each simulated action chunk is converted into a sequence of $k$ end-effector frame states in the robot base, augmented with the respective $k$ finger states. Let $\mathcal{K} = SE(3)\times\{0,1\}$ be the augmented keypoint space, with all possible Cartesian frame configurations and finger configurations.
\begin{equation}
    \forall t \in  \llbracket 1,m \rrbracket, \mathcal{T}^t_{RealChunk}=\left\{
\left(
\mathbf{F}^t_{\mathrm{eef}_i/\mathrm{real\_base}},
g_i^t 
\right)
\right\}_{i=1}^{k} \in \mathcal{K}^k.
\end{equation}

To achieve conversion, the relative Cartesian variations $\Delta \mathrm{eef}_i$ of a chunk are cumulatively used to transform the real robot initial configuration, $\mathbf{F}_{\mathrm{eef_1}}$. This generates a sequence of end-effector target frames in the real robot reference frame. The first target frame is obtained by applying  $\Delta \mathrm{eef}_1$ to the current real-world end-effector pose, while each subsequent keypoint is computed from the preceding target pose and the corresponding relative variation. 

We denote $f_\text{chunk2traj}$ the function that converts simulated action chunks into real-robot end-effector target state:
\begin{equation}
\begin{aligned}
f_\mathrm{chunk2traj}:\quad
SE(3)
\times \mathcal{A}^{k}
&\longrightarrow
\mathcal{K}^k
\\
\left(
\mathbf{F}^t_{\mathrm{eef_1}/\mathrm{real\_base}},
a^t_{SimChunk}
\right)
&\longmapsto
\mathcal{T}^t_{\mathrm{RealChunk}}.
\end{aligned}
\end{equation}

\textbf{Action Chunk Execution.} At each timestep, the processed trajectory target $\mathcal{T}^t_{\mathrm{RealChunk}}$ is used for real-world robot execution. 
More precisely, a MoveGroup path planner in joint space is used to transform the end-effector state target sequence into a joint control signal. In practice, Cartesian paths are interpolated with a 1 cm maximum step, and the arm is controlled by joint position controllers with fixed high gains. We observed that the planner hyperparameters and controller gains have a strong impact on both the quality of the transfer and its reproducibility. During a chunk execution, the robot arm pauses trajectory execution and
processes a state change at every index $i \in \llbracket 1,k-1 \rrbracket$
such that $g_i \neq g_{i-1}$. The command is executed through position control of the two symmetrically coupled finger joints, using predefined open and closed apertures and a maximum closing effort hyperparameter.

\subsection{Expert and Inference sim2real Deployment Protocols}
\label{subsec:open-loop}

\textbf{Expert data collection.} Real-world expert data collection consists of replaying several simulated experts in the real world.
One expert sample generated in simulation embodies information about the environment's initial configuration and the sequence of $m$ simulated actions $a_{SimChunk}$ to accomplish a given task fully. Real-world data collection includes an expert calibration phase that limits the gap introduced by the user between the expert simulated environment configuration and the real-world configuration. The real robot is used as a Cartesian pointer toward specific Cartesian target poses in order to guide the user in placing objects in the real robot expert scene.

All the sequences of expert chunks to be deployed begin at the same expert initial robot configuration $\mathbf{F}^{1}_{\mathrm{eef}_{1}/\mathrm{real\_base}              }$. Then execution of the $m$ expert chunks is open-loop, given that no real-world observation is needed. Concretely,  given that all the action chunks are sequential, the last end-effector key position relative to an action chunk $t$ serves as the real robot state input for the next action chunk process.
\begin{equation}
    \forall t \in \llbracket 1,m-1 \rrbracket, 
     \mathbf{F}^{t+1}_{\mathrm{eef}_{1}/\mathrm{real\_base}} = \mathbf{F}^{t}_{\mathrm{eef}_{k}/\mathrm{real\_base}}.
\end{equation}
% The open-loop expert replay follows the following strategy :
% \vspace{-0.3em}

% \vspace{-0.2em}

% \begin{center}
% \resizebox{0.85\columnwidth}{!}{%
% \begin{tikzpicture}[auto, node distance=1.0cm, >=latex']
%     \node [coordinate] (input) {};
%     \node [block, minimum height=0.5cm, right=1.2cm of input] (chunk2traj) {$f_{\mathrm{chunk2traj}}$};
%     \node [coordinate, right=1.5cm of chunk2traj] (output) {};

%     % Entrée
%     \draw [->] (input) -- 
%         node[font=\scriptsize] {$a^t_{\mathrm{SimChunk}}$} 
%         (chunk2traj);

%     % Sortie
%     \draw [->] (chunk2traj) -- 
%         node[font=\scriptsize] {$\mathcal{T}^t_{\mathrm{RealChunk}}$} 
%         (output);

%     % Continuité entre chunks
%     \draw [->, dashed]
%         (output)
%         -- ++(0,-0.8)
%         -| node[pos=0.25,below,font=\scriptsize]
%         {$\mathbf{F}^{t+1}_{\mathrm{eef}_0/\mathrm{real\_base}}
%         =
%         \mathbf{F}^{t}_{\mathrm{eef}_k/\mathrm{real\_base}}$}
%         (chunk2traj.south);

% \end{tikzpicture}%
% }
% \end{center}

% \vspace{-0.5em}

% \textbf{(3) Data labeling}
% If the real-robot accomplished the task, the sample is stored in the real-world expert demonstrations database. A real-robot expert sample stores expert chunks with its associated visual observations.

% The pipeline integrates user-friendly real-world data collection, relying on a loop based on steps (1)-(2)-(3), iterated for each sample. The automated protocol uses an interactive terminal to communicate with the user about the next actions to execute. 

\textbf{Closed-loop inference deployment.} For each inference deployment, the user chooses an initial environment configuration and a task to accomplish. The objective for the real robot is to accomplish the task by inferring actions based on visual observations. The main difference between expert deployment and inference deployment is that the $m$ inference action chunks are unknown at the beginning of the task. So the robot command is constructed online, chunk by chunk, using an online call to the inference model.

\begin{figure}[!t]
    \centering
        \includegraphics[width=\linewidth]{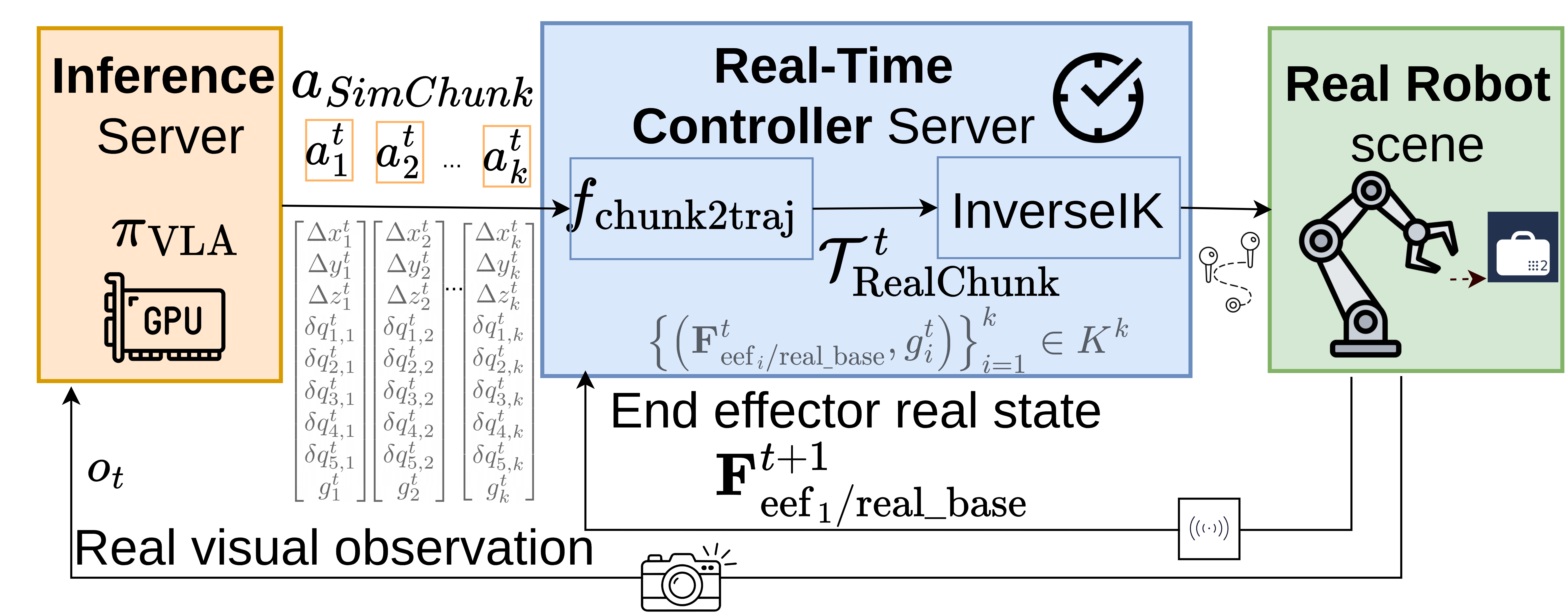}
   \caption{\textbf{Sim-to-real architecture} ROS 2 closed-loop pipeline with \textit{Inference Server} (\textcolor{orange}{orange}) and Real-Time \textit{Controller Server} (\textcolor{blue}{blue}) for real-robot deployment. }
\label{fig:ros2_architecture}
    \label{fig:action-server}
\end{figure}

There is a back-and-forth between $m$ robot action chunk predictions and robot command execution.
At each timestep $t \in \llbracket 1, m \rrbracket$ an action chunk $a^t_{SimChunk}$ is predicted based on the current visual observation $o_t$.
Given that each next action prediction is based on real-world visual feedback, this inference deployment is a closed-loop. Figure~\ref{fig:action-server} details the closed control loop. 
% \begin{figure}[!h]
% \centering
% \resizebox{\columnwidth}{!}{%
% \begin{tikzpicture}[auto, node distance=1.3cm,>=latex']
%     \node [block] (policy) {$\pi_\text{VLA}$};
%     \node [block, right=1.4cm of policy] (chunk2traj) {fchunk2traj};
%     \node [block, right=1.4cm of chunk2traj] (movegroup) {MoveGroupIK};
%     \node [block, right=1.4cm of movegroup] (robot) {Real robot};
 
%     \draw [->] (policy) -- node[font=\scriptsize] {$a^t_{\mathrm{SimChunk}}$} (chunk2traj);
%     \draw [->] (chunk2traj) -- node[font=\scriptsize] {$\mathcal{T}^t_{\mathrm{RealChunk}}$} (movegroup);
%     \draw [->] (movegroup) -- (robot);
 
%     \draw [->, dashed] (robot.south) -- ++(0,-1.0) -| node[pos=0.22,above,font=\scriptsize] {$\mathbf{F}^t_{\mathrm{eef}_0/\mathrm{real\_base}}$} (chunk2traj.south);
 
%     \draw [->] (robot.south) -- ++(0,-2.0) -| node[pos=0.08,above,font=\scriptsize] {$o_t$} (policy.south);
% \end{tikzpicture}%
% }
% \caption{Boucle fermée (\emph{closed-loop}) : à chaque épisode $t$, un chunk d'actions est inféré par $\pi_\text{VLA}$ à partir de l'observation $o_t$, converti en trajectoire réelle par \texttt{fchunk2traj}, puis exécuté sur le robot réel via \texttt{Move\_group}. L'état $\mathbf{F}^t_{\mathrm{eef}_0/\mathrm{real\_base}}$ atteint sert de point de départ au chunk suivant.}
% \label{fig:closed_loop}
% \end{figure}

% \textbf{(3) Success labelling}
% If the task was accomplished, the user labels with succes, else with failure.

% The pipeline includes an automated protocol to run repeated inference trials.

\subsection{ROS 2 architecture} 
\label{ROS2 sim2real communication interface}
A full task deployment relies on $t$ sequential executions of two ROS 2 custom \textit{Actions} \cite{ros2}. For each step $t$, first an \textit{Inference Action} tackles the computation of an action chunk $a_{SimChunk}^t$, then a \textit{Command Action} tackles its real-time deployment.

\textbf{{Inference Action.}}
First, the computation of an action chunk $a_{SimChunk}^t$ is done by a custom \textit{Inference Client/Server}. It involves asynchronous communication between the real-world observation $o_t$ and the trained model $\pi_\text{VLA}$.  For each step $t$, an \textit{Inference Action Client} sends a goal request containing $o_t$ to an \textit{Inference Action Server}. This server contains the $ \pi_\text {VLA} $. It computes a resulting action chunk $ a_{SimChunk}^t $ and sends it as a goal result.

\textbf{{Command Action.}}
Next, the real robot control to deploy $a_{SimChunk}^t$ is done by another custom \textit{Command Client/Server}. To do so, a \textit{Command Action Client} sends a goal request containing $a_{SimChunk}$ to a \textit{Command Action Server}. This server computes $\mathcal{T}^t_{\mathrm{RealChunk}}$ and controls the robot in real time using the process detailed in Section \ref{Chunking Learning Model Description}.

All the real-world experimental data is saved using \textit{Rosbag2}, which supports complex system recording in real time.

\section{Experiments}

\textbf{Expert real robot data collection.} We collected real-world data for VLA training with an action chunk size $k=8$ and $m=8$ chunks per trajectory i.e. $64$ actions per expert trajectory. We considered four language-conditioned combinations of two tasks (\textit{push right}, \textit{push left}) and two target objects (a \textit{red cube} and a \textit{blue cube}). We generate and deploy $50$ expert trajectories per task and compute metrics over the entire multi-task dataset composed of $200$ expert trajectories. To assess the sim-to-real gap of our pipeline, Table~\ref{tab:sim2real_metrics} reports the average end-effector (eef) path deviation and the Fréchet distance (which characterizes the maximal deviation) between the simulated Cartesian trajectories and their real-world executions. For a more granular analysis, we segment each trajectory into two distinct phases (based on a $z$-threshold): (1) the \textit{reaching} phase, where the robot approaches the cube, and (2) the \textit{contact} phase, where the robot manipulates the object. Finally, we report the overall task success rate for the real-world deployments.

\begin{table}[!h]
    \centering
    \resizebox{\columnwidth}{!}{%
    \begin{tabular}{l c c c}
        \toprule
        Metric & Reaching & Contact & Full \\
        \midrule
        Fréchet distance (cm) & $2.77 \pm 0.08$ & $3.30 \pm 0.05$ & $3.30 \pm 0.05$ \\
        Average eef path deviation (cm) & $0.90 \pm 0.03$ & $0.96 \pm 0.02$ & $0.93 \pm 0.02$ \\
        Task success rate & --- & --- & 100\% \\
        \bottomrule
    \end{tabular}%
    }
    \caption{Sim2Real metrics on cube pushing trajectories. Values are reported as mean $\pm$ 95\% confidence interval over 200 replayed trajectories.}
    \label{tab:sim2real_metrics}
\end{table}

Both metrics are computed between the end-effector target path in
simulation and the real end-effector measured path. These metrics are therefore
invariant to the temporal misalignment between the two executions. Across the full trajectory, the real replay deviates from the simulated path by less than one centimeter on average ($0.93$\,cm), with a worst-case deviation of $3.30$\,cm. The phase-wise breakdown shows where this gap originates: the average deviation is nearly identical in the reaching and contact phases ($0.90$ vs.\ $0.96$cm), indicating that the real controller consistently tracks the commanded path within the chosen $1$-cm interpolation tolerance set by the IK hyperparameter. In contrast, the Fréchet distance increases from $2.77$\,cm during reaching to $3.30$\,cm during contact. Since the Fréchet distance is a maximum over the path, its value on the full trajectory coincides with its value on the contact phase, which confirms that the largest deviations systematically occur while the robot interacts with the cube. In other words, the sim-to-real gap is not a global drift but consistent with contact-related effects, unmodeled in simulation (object mass, table friction), although tracking behavior near the table may also contribute. The $100\%$ success rate of the replayed expert trajectories indicates that this residual gap does not prevent the transferred demonstrations from completing the task during data collection.

\textbf{VLA evaluation.} As a proof of concept that the collected data can be used to train and deploy a chunk-based policy with the same stack, we train ORCHID~\cite{orchid} from scratch on the $200$ real-world demonstrations and evaluate it in closed-loop on $20$ trials ($5$ per task), with the target cube placed at a random position within the region where the task is feasible. The policy succeeds in $6/20$ trials, with a marked disparity across tasks ($3/5$, $2/5$, $1/5$ and $0/5$). This result should be read alongside the replay metrics: since the demonstrations themselves transfer with sub-centimeter deviation and complete the task, the limited success rate is unlikely to be explained by the physical gap alone. We rather attribute it to the small number of demonstrations per task and to the variability of the visual setting between collection and evaluation (lighting and shadows are not controlled), though the present experiments cannot separate these factors. We regard these experiments as a check of real-world deployment feasibility rather than a statistically powered evaluation of policy performance.

\section{Conclusion}
In this work, we present an open-source protocol that replays simulated expert trajectories on a real Franka FR3 to collect VLA-trainable demonstrations
without teleoperation, and reuses the same deployment stack for closed-loop policy evaluation. Because each real recording is paired with the simulated
trajectory that produced it, the protocol doubles as a policy-independent measurement of the sim-to-real gap. On $200$ cube-pushing replays, the real
end-effector path deviates from the simulated one by less than 1\,cm on average, and the largest deviations are confined to the contact phase.
Two limitations bound this result. First, the deviation is measured between a commanded simulated path and a measured real one, and thus aggregates the
physical mismatch with the tracking error of the real controller; isolating the former would require a proper identification of the real setup. Second,
open-loop replay is only meaningful for quasi-static tasks, and results are reported for a single robot, camera, and task family. The localization of the
gap in the contact phase nonetheless indicates where effort is best spent: identifying or randomizing object mass and table friction is likely to reduce
the residual gap more than any refinement of free-space motion. Lastly, a notable advantage of the ROS~2 pipeline is its modular architecture, which can be reused and customized to suit different models and sensors for sim-to-real integration.

\section*{Acknowledgments}

This work used IDRIS HPC resources under the allocation 2025-[AD011015740R1] made by GENCI. This work was partly funded by PostGenAI@Paris ANR-23-IACL-0007 France 2030 and the French National Research the OSTENSIVE project (ANR-24-CE33-6907-01).


\begin{thebibliography}{00}
\bibitem{octo} D. Ghosh, et al., “Octo: An Open-Source Generalist Robot Policy,” RSS XX, 2024.
\bibitem{openvla} M. J. Kim et al., “OpenVLA: An Open-Source Vision-Language-Action Model,” CoRL, 2024.
\bibitem{pi0} K. Black et al., “$\pi_0$: A Vision-Language-Action Flow Model for General Robot Control,” RSS XXI, 2025.
\bibitem{diffusion_policy} C. Chi et al., “Diffusion Policy: Visuomotor Policy Learning via Action Diffusion,” RSS XIX, 2023
\bibitem{droid} A. Khazatsky et al., “DROID: A Large-Scale In-The-Wild Robot Manipulation Dataset,” RSS XX, 2024.
\bibitem{lerobot} R. Cadene et al., “LeRobot: An Open-Source Library for End-to-End Robot Learning,” ICLR, 2026.
\bibitem{calvin} O. Mees et al., “CALVIN: A Benchmark for Language-Conditioned Policy Learning for Long-Horizon Robot Manipulation Tasks,” RA-L, 2022.
\bibitem{aljalbout}
E. Aljalbout et al., ``The Reality Gap in Robotics: Challenges, Solutions, and Best Practices,'' Oct. 23, 2025.
\bibitem{domain_rand} J. Tobin et al., “Domain Randomization for Transferring Deep Neural Networks from Simulation to the Real World,” IROS, 2017
\bibitem{frechet} Eiter, Thomas and Mannila, Heikki, "Computing Discrete Frechet Distance" TU Vienna, CD-TR 94/64, 1994.



\bibitem{pybullet} E. Coumans and Y. Bai, "PyBullet, a Python module for physics simulation for games, robotics and machine learning," \url{http://pybullet.org}, 2016--2021.







\bibitem{ros2} S. Macenski, T. Foote, B. Gerkey, C. Lalancette, and W. Woodall, "Robot Operating System 2: Design, architecture, and uses in the wild," Science Robotics, vol. 7, 2022
\bibitem{orchid} C. Grislain, et al., “Online Self-Training for Co-Adaptation in Hierarchical Diffusion Policies,” arXiv:2603.05291, 2026.

% \bibitem{b1} G. Eason, B. Noble, and I. N. Sneddon, ``On certain integrals of Lipschitz-Hankel type involving products of Bessel functions,'' Phil. Trans. Roy. Soc. London, vol. A247, pp. 529--551, April 1955.
% \bibitem{b2} J. Clerk Maxwell, A Treatise on Electricity and Magnetism, 3rd ed., vol. 2. Oxford: Clarendon, 1892, pp.68--73.
% \bibitem{b3} I. S. Jacobs and C. P. Bean, ``Fine particles, thin films and exchange anisotropy,'' in Magnetism, vol. III, G. T. Rado and H. Suhl, Eds. New York: Academic, 1963, pp. 271--350.
% \bibitem{b4} K. Elissa, ``Title of paper if known,'' unpublished.
% \bibitem{b5} R. Nicole, ``Title of paper with only first word capitalized,'' J. Name Stand. Abbrev., in press.
% \bibitem{b6} Y. Yorozu, M. Hirano, K. Oka, and Y. Tagawa, ``Electron spectroscopy studies on magneto-optical media and plastic substrate interface,'' IEEE Transl. J. Magn. Japan, vol. 2, pp. 740--741, August 1987 [Digests 9th Annual Conf. Magnetics Japan, p. 301, 1982].
% \bibitem{b7} M. Young, The Technical Writer's Handbook. Mill Valley, CA: University Science, 1989.
\end{thebibliography}
\end{document}